\def\year{2021}\relax
\documentclass[letterpaper]{article} % DO NOT CHANGE THIS
\usepackage{aaai21}  % DO NOT CHANGE THIS
\usepackage{times}  % DO NOT CHANGE THIS
\usepackage{helvet} % DO NOT CHANGE THIS
\usepackage{courier}  % DO NOT CHANGE THIS
\usepackage[hyphens]{url}  % DO NOT CHANGE THIS
\usepackage{graphicx} % DO NOT CHANGE THIS
\usepackage{natbib}  % DO NOT CHANGE THIS AND DO NOT ADD ANY OPTIONS TO IT
\usepackage{caption} % DO NOT CHANGE THIS AND DO NOT ADD ANY OPTIONS TO IT
\usepackage{epsfig}
\usepackage{subfig}
\usepackage{algorithm}
\usepackage{algorithmic}
\usepackage{multirow}

\usepackage{makecell}
\usepackage{enumitem}
\usepackage{multirow}
\usepackage{amsmath}
\usepackage[switch]{lineno}

\makeatletter \providecommand{\@LN}[2]{} \makeatother

\title{One for More: Selecting Generalizable Samples for Generalizable ReID Model}

\author{
Enwei Zhang\textsuperscript{\rm 1},
Xinyang Jiang\textsuperscript{\rm 1}\footnotemark[2]\thanks{Enwei Zhang and Xinyang Jiang contribute equally},
Hao Cheng\textsuperscript{\rm 1},
Ancong Wu\textsuperscript{\rm 2},
Fufu Yu \textsuperscript{\rm 1}, 
Ke Li \textsuperscript{\rm 1}, \\
\Large \textbf{Xiaowei Guo \textsuperscript{\rm 1}, 
Feng Zheng \textsuperscript{\rm 3}, 
Weishi Zheng \textsuperscript{\rm 2}, 
Xing Sun\textsuperscript{\rm 1}}\thanks{Corresponding Author (winfredsun@tencent.com, xinyangj@ zju.edu.cn)}

}

\affiliations{
    \textsuperscript{\rm 1} Tencent Youtu Lab, Shanghai, China \\
\textsuperscript{\rm 2} Sun Yat-sen University, Shenzhen, China \\
\textsuperscript{\rm 3} Southern University of Science and Technology, Shenzhen, China \\

}

\begin{document}

\maketitle

\begin{abstract}
Current training objectives of existing person Re-IDentification (ReID) models only ensure that the loss of the model decreases on selected training batch, with no regards to the performance on samples outside the batch. It will inevitably cause the model to over-fit the data in the dominant position (e.g., head data in imbalanced class, easy samples or noisy samples). %We call the sample that updates the model towards generalizing on more data a generalizable sample. 
The latest resampling methods address the issue by designing specific criterion to select specific samples that trains the model generalize more on certain type of data (e.g., hard samples, tail data),  which is not adaptive to the inconsistent real world ReID data distributions. 
Therefore, instead of simply presuming on what samples are generalizable, this paper proposes a one-for-more training objective that directly takes the generalization ability of selected samples as a loss function and learn a sampler to automatically select generalizable samples. More importantly, our proposed one-for-more based sampler can be seamlessly integrated into the ReID training framework which is able to simultaneously train ReID models and the sampler in an end-to-end fashion. The experimental results show that our method can effectively improve the ReID model training and boost the performance of ReID models.
 
%\keywords{}
\end{abstract}

%%%%%%%%% BODY TEXT
\section{Introduction}
% what is person-reid, the problem with person-reid: data distribution (hard sample, data imbalance and noise)
Person re-identification (ReID) aims at recognizing pedestrians across non-overlapping camera views, which increasingly draws attention due to its wide applications in surveillance, tracking, smart retail, etc \cite{zheng2015scalable,Zhao_2017_ICCV,Sun_2018_ECCV}. As deep learning prevails, the CNN based ReID methods progress rapidly and achieve impressive performance on benchmark datasets. However, ReID remains a challenging problem due to the view variance, domain changes, partial occlusion and many other factors. Most of the existing ReID methods solve the these issues by designing new network structures or innovative loss functions \cite{Sun_2018_ECCV,wang2018learning}, but very few pay attention to the nature of training data distribution. 

\begin{figure}[!t]
\centering
\includegraphics[width=0.5\textwidth]{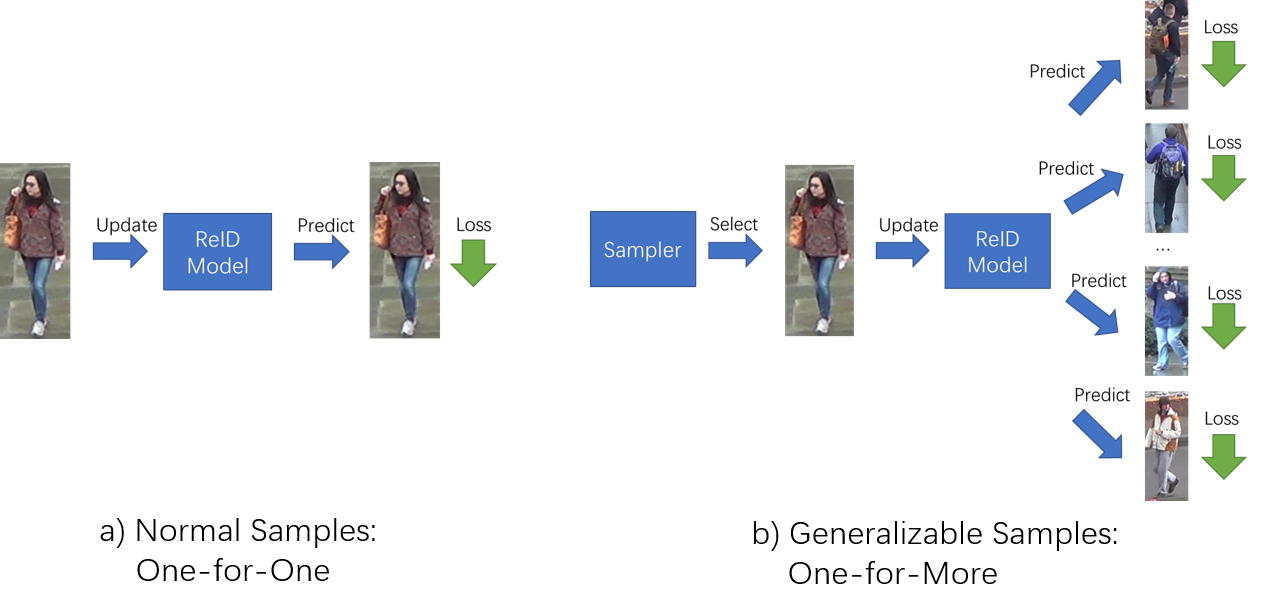} 
\caption{Comparison between one-for-one objective and one-for-more objective. } 
\label{fig_objectives}
\end{figure}

%current method
%As shown in Figure \ref{fig_sample_objective} a)

In standard ReID training task, at each iteration, after updating the model with a batch of selected samples, a normal training objective only makes sure the loss of these sample decreases, but has no guarantee that the loss of the samples outside the batch will also decreases, as shown in Figure \ref{fig_objectives} a). 
%We call the standard training objectives as one-for-one objectives. 
We call this type of objectives one-for-one objectives because the samples are selected for improving model performance on themselves, with no regards to performance on the samples outside the batch. As shown in \ref{fig_objectives} b), after updating model with the selected sample, if the loss of more samples decreases other than the sample itself, we call this sample a more generalizable sample. Intuitively, training with generalizable samples gives a model with more generalization ability.
%Since one-for-one objectives do not ensure the model to generalize on all samples at each update iteration, when a certain type of samples take a dominant place (e.g., imbalanced class, easy samples or noisy samples) , it will easily cause the model to over-fit. 

Real world ReID data distribution is usually highly imbalanced with notable amount of hard samples (e.g., view/pose change, occlusion, camera resolution etc) and noisy data, and one-for-one objective could potentially cause models over-fitting on a dominate part of the data. 
To avoid over-fitting, methods including re-sampling and re-weighting propose different criteria to select samples to train the model to generalize more on certain type of data. For example, following are some typical selection criteria for state-of-the-art ReID methods: 
\begin{itemize}[leftmargin=*]
\item \textbf{Select Easy Sample}
Methods like curriculum learning \cite{wang2018mancs}\cite{guo2018curriculumnet} pre-define noisy labels as samples with low training loss or high prediction confidence and proposes to emphasize on these samples to prevent model over-fitting on noisy labels. 
\item \textbf{Select Hard Sample}
Methods like hard sample mining \cite{HermansBeyer2017Arxiv} and focal loss \cite{lin2017focal} define hard samples as samples with large training loss or low prediction confidence, and proposes to emphasize on these samples to help model generalize better on samples difficult to recognize. 
\item \textbf{Learn Easy Sampler}  Methods like self-paced Learning \cite{zhou2018deep} and abstention \cite{thulasidasan2019combating} do not directly use a selection criteria to select samples. 
Instead, the criteria becomes an objective function to train an automatic data sampler. For example, Abstention trains an easy data sampler to select samples with the minimum training loss. %use the training loss as a supervised signal to train a easy data sampler that automatically select sample with low losses.  
\item \textbf{Learn Hard Sampler}
Similarly to easy data sampler, methods like DE-DSP\cite{duan2019deep} train an automatic hard data sampler by maximizing the the training loss of selected samples. 
%Meta-weight Net \cite{shu2019meta} uses meta-learning method to train sampler that select samples based on training loss. 
%Although DE-DSP does not directly apply manual rules to compute sample importance, the sampler is still trained to approximate a rule to select samples with large losses, with a upper-bound of hard sample mining. 

%Secondly, besides the property of data distribution, empirical study shows \cite{bengio2009curriculum,guo2018curriculumnet} that different learning stages also affect the choice of sampling policy, such as the easy-to-hard learning schedule manually defined in curriculum learning.
\end{itemize}

\begin{figure}[!t]
\centering
\includegraphics[width=0.5\textwidth]{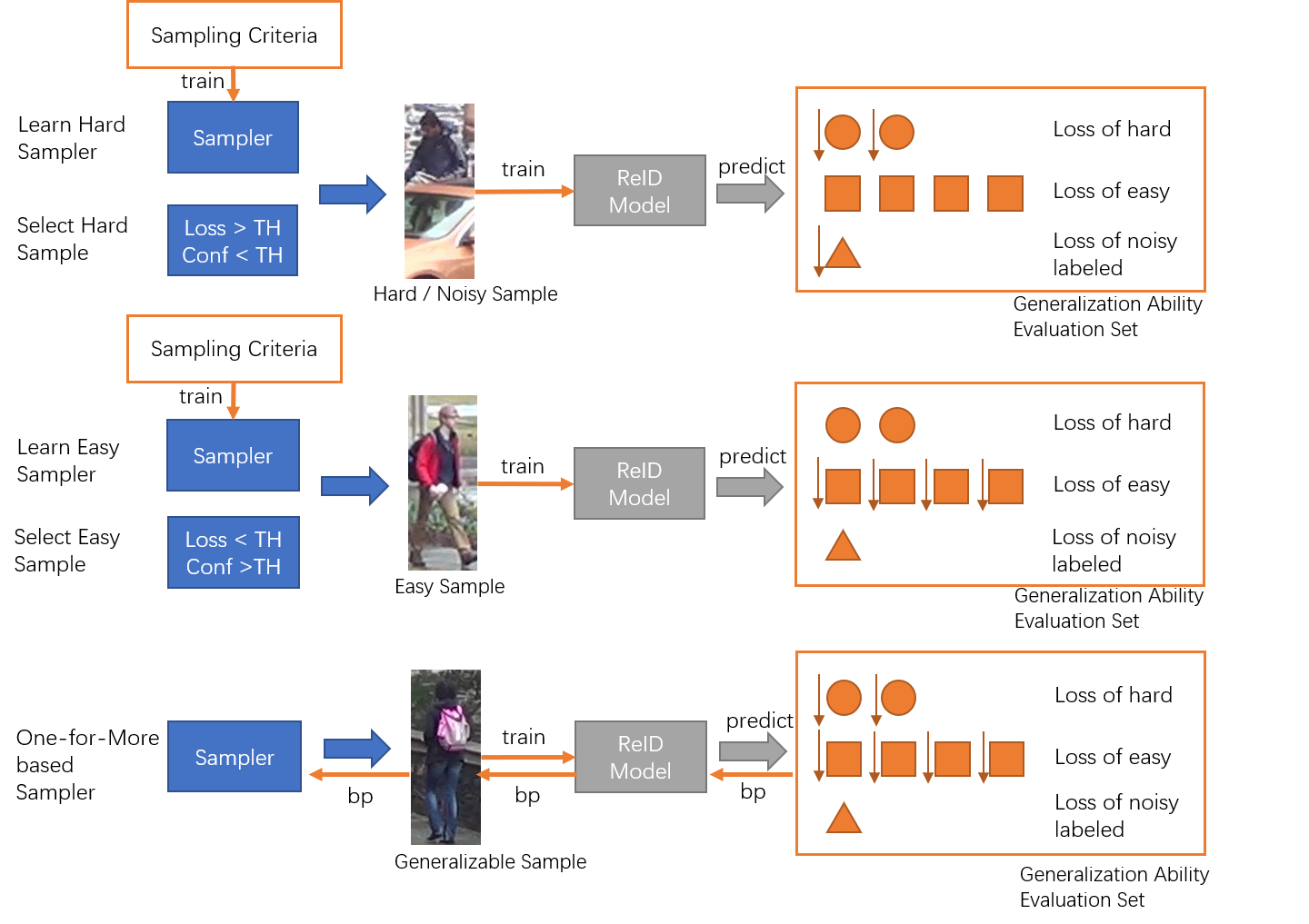} 
\caption{Comparisons of our methods with different sampling methods. } 
\label{fig_methods_compare}
\end{figure}
Besides above-mentioned methods, there are also other types of selection criteria making model generalize on different types of samples, such as tail data (up-sampling or over-sampling), image with certain attributes \cite{wu2018person}. 

All the above-mentioned methods are based on selection criterion that  selects specific samples to train model to generalize on certain type of data (e.g. hard samples, tail data or clean data). Thus, these methods are usually not adaptable when training on another type of data distribution. For example, as shown in Figure  \ref{fig_methods_compare}, hard sampling causes over-fitting on noisy labeled data which also has large training loss, and easy sampling causes over-fitting on easy samples that takes dominate place in training set. 
As a result, instead of giving unreliable presumption of what is good criterion for selecting generalizable sample, we propose a new objective and criterion that directly takes sample's ability to make model generalize on more data as a loss function, called one-for-more objective. As shown in last row of Figure \ref{fig_methods_compare}. A batch of samples are first selected to update the ReID model. Then, another evaluation set of samples outside the training batch is sampled, and the generalization ability of the selected training batch is evaluated by computing the updated model's loss on this set. Thus, the data sampler is trained by minimizing the training loss outside the selected training batch, which is to  back-propagate the training loss. The detail implementation will be introduced in section 3. 
% learn a sampler that is able to select samples to update model towards generalizing on all data. 
%Thus, as shown in \ref{fig_objectives} b) we propose a novel one-for-more objective which enforces the sampler to select samples that makes the loss of the ReID model to decrease over all images in the training set. 
%As shown in \ref{fig_methods_compare}, compared to other sampling methods that could potentially over-fits on certain type of samples, the goal of the one-for-more based sampling is to make model generalizes on all samples. 
%Furthermore, As shown in \ref{fig_methods_compare}, compared to learning based sampling methods that uses pre-defined criteria as supervised signal,  % pre-defines supervised signals based on  , 
%our method directly back propagates the training losses of all samples to the data sampler. 
%Furthermore, we argue that the information inside each individual sample is not enough for a sampler to evaluate the importance of the sample. 

%\begin{figure}[!t]
%\centering
%\includegraphics[width=0.8\textwidth]{figures/motivation1.png} 
%\caption{The illustration of one-for-more Objective for automatic data sampler training. } 
%\label{fig_motivation}
%\end{figure}

%\ancong{/*** ancong: some details of the framework can be more concise in the next two paragraphs ***/}
%Figure \ref{fig_motivation} is a illustration of the data-flow of our proposed one-for-more objective to train the automatic data sampler.  

The one-for-more sampler training process is integrated into the ReID model training process, forming an end-to-end framework, which at each iteration, simultaneously updates the data sampler and updates the ReID model with samples selected by the sampler. 
To make the data sampler more suitable for ReID, besides sampling individual images, we propose a pair-wise data sampler that considers correlation between images and support pair-wise sample selection for contrastive loss. % individual image feature and the image's correlation with other samples.

%our solution
In conclusion, to avoid over-fitting caused by standard one-for-one objective in a complex real world data distribution, this paper proposes an one-for-more objective based end-to-end framework that simultaneously trains ReID model with selected samples and the data sampler. Instead of optimizing sample itself, our method is able to select generalizable samples that update the ReID model towards generalizing on samples outside the training batch.

\section{Related Work}
As mentioned in the last section, to prevent over-fitting on small sub-set of the training data caused by one-for-one objectives, over the years, many sampling based works have been proposed to adapt different types of data distribution. 

\textbf{Imbalance Identities}. Re-sampling/re-weighting \cite{he2009learning,chawla2002smote,shrivastava2016training} based methods are one of the most widely used strategy to tackle imbalance problem including up-sampling the minority samples or down-sampling the majority samples. Other works mainly focus on adjusting the model or loss function. For example, \cite{zhang2017range} %\ancong{/*** ancong: It would be better to use the names of authors or method as the subject of the sentence ***/} 
solves the imbalance problem between relevant pairs and irrelevant pairs of samples by introducing a range loss that defines the sample relevance at identity level. \cite{zhong2019unequal} propose to treats head data and tail data in different way , and apply two different sets of loss functions. \cite{liu2019large} propose an external memory structure to help learn better features on tail samples, but their method is for classification task, which is not compatible with recognition and retrieval. 

%\ancong{/*** ancong: For each type of related methods, we should discuss the difference between them and our approach and emphasize what they cannot solve but we can ***/}

\textbf{Hard Sample Mining}.   
One of the goals of re-sampling and re-weighting methods is to locate and emphasize the learning on samples hard to converge (e.g., samples with small inter-class variation and large intra-class variation). Adaboost \cite{freund1997decision} is a widely used machine learning algorithm, which iteratively trains new models on hard samples found by old models. Online hard negative mining \cite{shrivastava2016training} and focal loss \cite{lin2017focal} propose loss based hard sample mining method that selects hard training samples for object detection. \cite{HermansBeyer2017Arxiv} proposes a easy-to-implement online hard mining method for triplet loss and achieves great performance improvement in person re-identification. \cite{schroff2015facenet} propose to only select negative samples that further away from the anchor than the positive exemplar, but the distance to anchor still close to the anchor-positive distance (i.e. semi-hard samples). 
DE-DSP\cite{duan2019deep} and Meta-weight Net \cite{shu2019meta,pmlr-v80-ren18a}  propose to learn re-weighting functions that automatically assign importance weight to training samples, which is somewhat similar to our method. Compared to these works, our methods propose a one-for-more objective which able to select samples that update model toward generalize on the entire training set. 

\textbf{Noise Resist}. 
Most of the person re-identification datasets contain a small portion of noisy labels, which requires the learning algorithm to resist certain amount of noises. Many studies propose to strengthen the resistance toward noisy labels by customized sampling strategy  \cite{kumar2010self}, network structure \cite{Yu_2019_ICCV}, data cleaning \cite{tanaka2018joint} or data augmentation \cite{JMLR:v15:srivastava14a,zhang2018mixup}. The experiments will show that our method is robust to noisy labels as long as the noisy level is below $20\%$ (consistent with most of the ReID data distribution). %, our method achieves comparable methods with the state-of-the-art noise resist method. 
%self-paced learning \cite{kumar2010self} proposes to train models on reliable samples first and gradually add samples that potentially contain noisy labels at the later stage of the training. \cite{hu2019noise} follows the similar multi-stage schema and proposes an easy to hard training framework for face recognition. \cite{Yu_2019_ICCV} propose a novel CNN structure for person re-identification, where the features of noisy samples under the same identity are modeled as a Gaussian distribution, so that the outliers (high variation samples) are separated from the clean data. \cite{tanaka2018joint} proposes a joint training framework that iteratively updates model and correct noisy labels during training. Although not explicitly dealing with noisy labels, regularization methods are also an efficient way to prevent deep model from overfitting to noisy labels, such as weight decay and dropout \cite{JMLR:v15:srivastava14a}. Data augmentation is one of the ways to directly alter the distribution of training set by making minor transformation on training samples, which implicitly makes models generalize better on noisy or hard samples. For example, mix-up \cite{zhang2018mixup} uses a random convex combination of images and labels as training samples. 

\section{The Proposed Method}
\begin{figure*}[!t]
\centering
\includegraphics[width=0.7\textwidth]{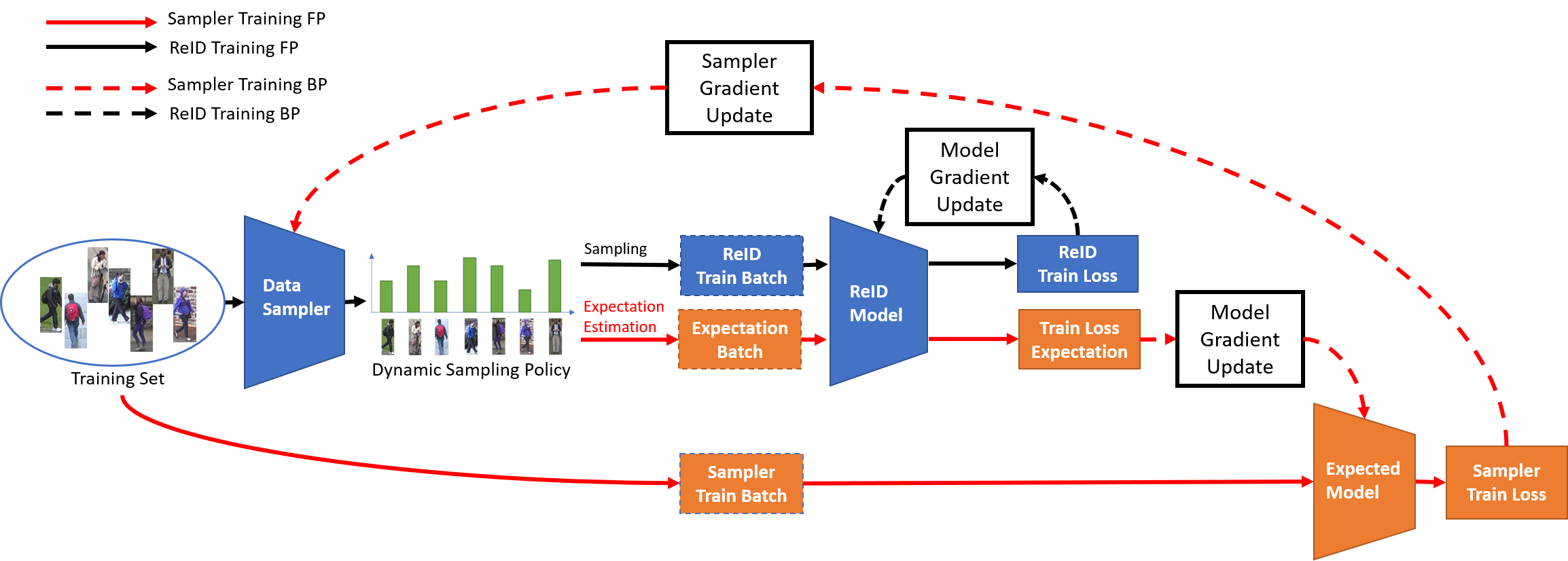} 
\caption{The detailed workflow of ReID method with proposed dynamic sampling policy learning. } 
%\ancong{/*** ancong: more color can be used for different blocks, and different processes can be separated by dashed lines. Is the validation batch shown in this figure? Why the shapes of train input expectation and updated model expectation are different? It is better to show the notations in the formulations in this figure. ***/}
\label{fig_pipeline}
\end{figure*}

Our method trains two neural network models, namely the ReID model and the data sampler. The data sampler is responsible for outputting a sampling policy to select samples from the training set at each training iteration and then the ReID model is updated with the selected training batch. As shown in \ref{fig_pipeline}, there are two learning branches in our proposed method, namely the ReID model learning process marked with black arrows and the data sampler learning process marked with red arrow, which are jointly trained in an end-to-end fashion. 
%\ancong{/*** ancong: distribution learning may be misleading as it was already defined for other problem ***/} marked with red arrows. %As shown in Figure \ref{fig_pipeline}, the original data distribution (i.e. a uniform distribution over the training set) is transformed into a new distribution over the training set. 

The ReID model learning process is a typical deep learning schema. At each iteration, a mini-batch is drawn from training set with the sampling policy (i.e. sampling probability of each sample) generated by data sampler and fed into the ReID model to obtain the training loss.  The gradient of the training loss is then back propagated to update the model weights. 

The data sampler learning process involves two model update processes in one iteration. First, a sample batch selected by the sampler is used to update the ReID model. Then, another sampler training batch is drawn from the original training set distribution to compute the one-for-more sampler training loss with the updated ReID model. Finally, one-for-more loss is  back propagated all the way through the ReID update operation to the data sampler.
Since data sampling process is non-derivable, instead of directly sampling a batch, we approximate an expectation of the training batch sampled with current data sampler.    %Then we compute the convergence status of the expected new ReID model by computing the training loss with a data sampler training batch. Finally, the training loss of the data sampler training batch is back-propagated to update the data sampler. 

We elaborate on the detailed implementation of the two learning branches in the following sections.

\subsection{Background}

For a person re-identification problem, given any image denoted as $x$, we need to train a ReID model $h$ to predict its corresponding identity $y$. 

%Assuming probability distribution of training dataset over $X$ and $Y$ is $q(x, y)$, the objective function is to minimize the risk associated with hypothesis $h$, which is defined as the expectation of the loss function:

%%\begin{equation}
%h^*=\arg\min_{h}R(h), 
%\end{equation}
%where, 
%\begin{equation}
%R(h) = E\big[L\big(h(x, w), y\big)\big] = \int L\big(h(x, w), y\big)d q(x, y), 
%\end{equation}
%where, $w$ is the model parameter of ReID model. 

%However, the data distribution of training set is unknown, and what we have is a set of $N$ training examples $(x_1, y_1), ..., (x_n, y_n)$ drawn from the original data distribution. We use the average of the loss function of this samples to approximate the risk, called the empirical risk:
%\begin{equation}
%R_{emp}(h) = \frac{1}{N}\sum_{i=1}^n L\big(h(x_i, w), y_i\big). 
%\end{equation}

Deep learning based ReID model uses mini-batch stochastic gradient descent algorithm to solve the ReID loss function. At each iteration, a small sub-set of samples are drawn from the training set for model update. Denoting the sampling policy as $\hat{P}$, at each iteration, the model weights is updated with the mini-batch drawn with the sampling policy: 
\begin{equation}
w^{(t + 1)} = w^{(t)} - \beta \nabla_{w^{(t)}} \frac{1}{K}\sum_{x_i\sim \hat{P}}L\big(h(x_i, w^{(t)}), y_i\big), 
\label{eq_reid_weight_update}
\end{equation}
where, $w^{(t)}$ is the parameters of the ReID model at $t$-th iteration, $K$ is the size of the mini-batch, and $L$ is the ReID Loss function. 
The goal of our method and other re-sampling methods is to find an optimal sampling policy that selects genalizable samples for ReID model update. 

%In this paper we use a ResNet-50 model as the ReID feature extractor and Cross-entropy and Triplet Loss as the loss function. For triplet loss, to make sure samples in triplet has the correct labels, we independently draw anchors, positive samples and negative samples from the generated distribution. 

\subsection{One-for-more Objective for Data Sampler}
\subsubsection{Data Sampler Modeling}
%In this sub-section, we introduce our method to learn the training set distribution $\hat{P}(x)$. 

For most of the existing re-sampling methods, sampling policy $\hat{P}$ is not automatically learned, but a predefined fixed strategy. 
%Given a training set with $N$ samples, the probability of selecting any image $x$ in a training set is:
%\begin{equation}
%    \hat{P}(x) = \frac{P(x)I(x)}{\sum_{i=1}^N{P(x_i)}{I(x_i)}} , 
%    \label{eq_2stage_sampling}
%\end{equation}
%where $P(\cdot)$ is a uniform sampling probability to draw samples from the original data distribution; $I(\cdot)$ is a indication function defining different re-sampling strategies; 
For example, online hard sample mining defines a policy that first sampling mini-batch from training data from a uniform probability and then selects triplet samples with lowest similarity in from this batch. 
%$I(\cdot)$ as:
%\begin{equation}
%    I(x) = \left\{
%    \begin{aligned}
%    &1, x \in A  \\ 
%    &0, x \notin A
%    \end{aligned}
%    \right.
%\end{equation}
%where $A$ is a set containing $k$ samples with largest $L(h(x, w), y)$. 

Instead of a predefined strategy, we propose to learn a data sampler that predicts the probability $\hat{P}(x| h)$ to select sample $x$ given current ReID model $h$. By choosing Multi-layer perception as an energy function, the probability for sampler to select an image $x$ is modeled as:
\begin{equation}
    \hat{P}(x| h) =  \frac{e^{MLP\big(h(x, w)), \theta\big)}}{\sum_{i = 1}^N e^{MLP\big(h(x_i, w)), \theta\big)}}, 
\end{equation}
where $\theta$ is the set of parameters of probability function $\hat{P}$ and $N$ is the size of the dataset. 

The above-mentioned sampling policy only considers visual information of single images and is not suitable for sampling image pairs or triplets. We further extend the proposed data sampler to predict sampling policy based on mutual information of sample pairs. We model the probability to select any image $x$ given an anchor image $x_a$ and ReID model $h$ is modeled as follows:  
\begin{equation}
    \hat{P}(x|x_a, h) =  \frac{e^{MLP\big([h(x, w), h(x_a, w)],  \theta\big)}}{\sum_{i = 1}^N e^{MLP\big([h(x_i, w), h(x_a, w)], \theta\big)}}, 
\end{equation}
where $[h(x_i, w), h(x_a, w)]$ is the concatenation of to feature vector.  %and the sampling probability for a image pair $(x_0, x_1)$ is:
%\begin{equation}
% \hat{P}(x_0, x_1|, h) = \hat{P}(x_0, h)\hat{P}(x_1|x_0, h)
%\end{equation}

In conclusion, we sample individual image batches to optimize cross entropy based loss with single image sampler $\hat{P}(x|h)$. For triplet loss, we first sample anchors with single image sampler, and then given an anchor image, we sample corresponding positive and negative sample with pair-wise sampler ${\hat{P}(x|x_a, h)}$ to form a image triplet.

\subsubsection{Data Sampler Optimization}
%The goal is to find an optimal distribution that after update the model with a mini-batch drawn from the distribution, the expected model achieves minimum loss on a validation batch drawn from the original distribution containing $M$ samples $(x'_1, y'_1), .... (x'_m, y'_m)$:
As shown in Figure \ref{fig_pipeline}, the optimization of data sampler involves two model  update processes. The model is first updated with a batch of samples selected by the trained sampler, and then the objective function for the data sampler is to minimize the ReID loss of the updated model over another set of data, i.e. the one-for-more loss. 

In detail, since the process of sampling data from a distribution is non-derivable, we use a expectation approximation to replace the actual data sampling. 
At $t$-th iteration, we approximate the expectation of the updated model's weights, given current sampling policy $\hat{P}$, denoted as %,  what the updated model will be after one learning step (i.e. compute the expectation of the ReID model weight
$E_{\hat{P}}[w^{(t+1)}]$. Then the optimal policy $\hat{P}$ is obtained  by minimizing the updated model's training loss over all $N$ training samples:
\begin{equation}
\begin{aligned}
    \hat{P}= &\arg\min_{\hat{P}}R'(\hat{P}) \\
    &=\arg\min_{\hat{P}} \sum_i^N L\big(h(x_i, E_{\hat{P}}[w^{(t+1)}]), y_i\big). 
\end{aligned}
\label{fig_distribution_objective}
\end{equation}
%where  is the expectation of the weights of updated ReID model $h$ and 
%Following SGD, at each iteration a sampler training batch with size $M$ is drawn to update the data sampler's parameter. 

Next we introduce how to obtain the above-mentioned updated model weight expectation $E_{\hat{P}}[w^{(t+1)}]$. Assuming at each learning iteration, we draw $n$ single images $x_i$ from training set containing $N$ images. Then the expected number of sample $x_i$ occurs in the drawn mini-batch is $n\hat{P}(x_i|h)$, and then given the generated sampling policy $\hat{P}$, the expectation of the ReID model's loss on the drawn batch is: 
\begin{equation}
\begin{aligned}
E_{\hat{P}}[L\big(h(x), y\big)] = \sum_{i=1}^N \hat{P}(x_i) L\big(h(x_i, w^{(t)}), y_i\big), 
\label{eq_loss_expectation}
\end{aligned}
\end{equation}
Similarly, the triple loss of data sampled by the image pair sampler is formulated as:
\begin{equation}
\begin{aligned}
   &E_{\hat{P}}[L(h(x), y)] = \sum \hat{P}(x_a, x_p, x_n| h)L_{tri} \\ 
   & = \sum  \hat{P}(x_a| h)\hat{P}(x_p|x_a, h)\hat{P}(x_n|x_a, h)L_{tri}, 
\end{aligned}
\end{equation}
where $L_{tri}$ is short for:
\begin{equation}
\begin{aligned}
\tiny
L_{tri} &= max\Big(d\big(h(x_a, w^{(t)}), h(x_p, w^{(t)})\big) \\ 
& - d\big(h(x_a, w^{(t)}), h(x_n, w^{(t)})\big) + m, 0\Big), 
\end{aligned}
\end{equation}
where $x_a$ is the anchor image; $x_p$ is the positive sample which has the same identity with anchor image; $x_n$ is the negative sample of the triplet; $m$ is the margin value. 
Following the mini-batch based SGD, it is infeasible to obtain loss expectation based on the entire training set. We draw a subset containing $K$ samples to approximate the loss expectation, where the values of the samples' selecting probability is normalized to sum to $1$. 

%\begin{figure}[t]
%\centering
%\includegraphics[width=0.5\textwidth]{figures/sampling_triplet.png} 
%\caption{The proposed data sampler adapts to both single image sampling for cross-entropy based loss and image pair sampling for triplet loss. } 
%\label{fig_sampling_triplet}
%\end{figure}

Due to the page limit, in rest of this section, we take single image sampler $\hat{P}(x|h)$ as an example to introduce the proposed sampling policy learning methods. Now that we obtain the train loss expectation, we use a common SGD method to update the ReID model. By taking the derivative of the loss expectation, we obtain the expected gradient of the loss with respect to the model weight, noted that the expectation is approximated on a batch containing $K$ samples:
\begin{equation}
\small
E_{\hat{P}}[\nabla_{w^{(t)}}L\big(h(x), y\big)] = \sum_{i = 1}^K\hat{P}(x_i|h)\nabla_{w^{(t)}}L\big(h(x_i, w^{(t)}), y_i\big). 
\end{equation}
As a result, following SGD, the expected model weight is updated as:
\begin{equation}
\small
E_{\hat{P}}[w^{(t+1)}] = w^{(t)} - \beta \sum_{i = 1}^K\hat{P}(x_i|h)\nabla_{w^{(t)}}L\big(h(x_i, w^{(t)}), y_i\big). 
\label{eq_weight_expectation}
\end{equation}

Finally we evaluate the training loss of the updated model  over all samples in the training set. %expectation under current policy $\hat{P}(x|h)$, 
Given a sampler training batch with $M$ samples is randomly sampled, we substitute  Eq.\eqref{eq_weight_expectation} to the optimization objective in Eq.\eqref{fig_distribution_objective} and obtain the training loss expectation over $M$ samples, and compute the gradient of this loss function with respect to data sampler's weight $\theta$. Given the gradient,  the weights of the data sampler are updated as follows:
\begin{equation}
    \theta^{(t + 1)} = \theta^{(t)} - \alpha \sum_{i=1}^M \nabla_\theta L\Big(h\big(x_i, E_{\hat{P}}(w^{(t+1)})\big), y_i\Big)
\label{eq_transfer_weight_update}
\end{equation}

%As shown in Eq.\ref{eq_transfer_weight_update}, the training of the data sampler involves the computation of secondary derivatives, which we found in the experiments is extremely slow in the training process and very hard to converge. As a result, in order to speed up the training, we propose another simplified objective function that serves the same motivation. Instead of learning a data sampler that outputs each sample's probability of getting selected, we first train a regressor that predicts sample generalizing ability. In other word, the loss changes on the sampler training data is used as the label of the regressor. 

\subsection{ReID Model Training with Automatic Data Sampler}

Our training framework uses both cross entropy loss and triplet loss, which is two types of loss function achieves sate-of-the-art performance. 
%As shown in Figure \ref{fig_sampling_triplet}, 
Our framework learns a single image sampler $\hat{P}(x|h)$ to sample individual images and learns an image pairs sampler $\hat{P}(x|x_a, h)$ to sample image pairs. We first use single image sampler to draw anchor images, and then use image pair sampler $\hat{P}(x|x_a, h)$ to sample positive and negative image pairs given the anchor image.  
%To simultaneously optimize cross entropy loss and triplet loss,
%For cross entropy loss, the single image sampler is used to individually sample images to form a mini-batch. 

Directly obtaining the probability of image pairs in $N^2$ space is infeasible during training. Thus, given the anchor image $x_a$, we design a semi-global sampling process to approximate the above-mentioned pair-wise sampling. First we use single image sampler $\hat{P}(x|h)$ to sample a positive candidate set and a negative candidate set globally from the entire training set, and then the positive image and negative images are sampled from the candidate set with $\hat{P}(x|x_a, h)$ respectively.  %the image pair sampler $\hat{P}(x|x_0, h)$ is used to sample positive pairs. As the negative sampler  and negative pairs corresponding to the anchor images. 
We summarize the entire pipeline of our proposed training framework step-by-step in Algorithm \ref{alg_1}. 

%where $h^{(t+1)}$ is the ReID model updated one iteration with the expected data sampler:  
%\begin{equation}
%    h^{(t+1)}(x') = h(x', ), y_i))
%\end{equation}

%After updating the data sampler, the probability to be selected of all samples needs to be obtained with the new sampler. To accelerate the distribution updating in each iteration and make the predicted probability more robust and smooth over the time, we approximate the global sampling probability by updating samples in the sampler training batch at each iteration in a moving-average fashion: 
%\ancong{/*** ancong: use consistent notation with the algorithm. Why updating P for another randomly sampled batch $x''$? ***/}
%\begin{equation}
%    \hat{P}^{(t+1)}(x) =(1 - \mu)\hat{P}^{(t)}(x) + \mu \frac{MLP(h(x, h))}{\sum_{i = %1}^KMLP(h(x_i, h))}
%\label{eq_probability_update}
%\end{equation}

%\ancong{/*** ancong: it is not explained in the methodology how the challenges of long tail distribution and visual ambiguity are solved by the dynamic policy. More analysis is needed to link challenges with the approach. ***/}

\begin{algorithm}
\caption{Person Re-Identification with Dynamic Data Sampler Learning}
\label{alg_1}
\begin{algorithmic}
\REQUIRE
Training data $D_{tr}$
\ENSURE
ReID model parameter $w^{(T)}$
\STATE Initialize  $w^{(0)}$ and $\theta^{(0)}$.  
\FOR{$t = 0$ \textbf{to} $T-1$}
\STATE Sample anchor images $x_a \sim \hat{P}(x|h)$
\FOR{each $x_a$}
\STATE Sample positive candidate set $X_p$
\STATE Sample negative candidate set $X_n$
\STATE Sample positive pair $x_p$ from $X_p$ with $\hat{P}(x_n|x_a, h)$
\STATE Sample negative pair $x_n$ from $X_n$ with $\hat{P}(x_n|x_a, h)$
\STATE Add $(x_a, x_p, x_n)$ into ReID training batch
\ENDFOR
\STATE Update ReID model weights $w^{(t+1)}$ with Eq. \eqref{eq_reid_weight_update} using ReID training batch 
\STATE Sample expectation evaluation batch
\STATE Obtain weight expectation of the data sampler with expectation evaluation batch by  Eq. \eqref{eq_weight_expectation}
\STATE Sample data sampler training batch 
\STATE Update weights of data sampler $\hat{P}$ with sampler training batch by Eq. \eqref{eq_transfer_weight_update}
\ENDFOR

\end{algorithmic}
\end{algorithm}

\section{Experiments}
\subsection{Dataset and Evaluation Metrics}
Our experiments are conducted on three widely used ReID benchmark datasets. 

\noindent \textbf{Market-1501} 
dataset contains 32,668 person images of 1,501 identities captured by six cameras. Training set is composed of 12,936 images of 751 identities while testing data is composed of the other images of 750 identities. In addition, 2,793 distractors also exist in testing data. 

\noindent \textbf{MSMT-17} 
dataset contains 124,068 person images of 4,101 identities captured by 15 cameras (12 outdoor, 3 indoor). Training set is composed of 30,248 images of 1,041 identities while testing data is composed of the other images of 3060 identities.  

%\vspace{-0.2cm}
\noindent \textbf{DukeMTMC-reID} 
dataset contains 36,411 person images of 1,404 identities captured by eight cameras. They are randomly divided, with 702 identities as the training set and the remaining 702 identities as the testing set. In the testing set, For each ID in each camera, one image is picked for the query set while the rest remain for the gallery set. 

\noindent \textbf{Evaluation Metrics.}
Two widely used evaluation metrics including mean average precision (mAP) and matching accuracy (Rank-1/Rank-5) are adopted in our experiments.

\subsection{Implementation Details}
The input image size is set to $256 \times 128$. The ResNet-50 model with the pretrained parameters on ImageNet is chosen as the backbone network. Common data augmentation include horizontal flipping, random cropping, padding, random erasing (with a probability of $0.5$) are used. We adopt Adam optimizer to train our model and set weight decay $5 \times 10^{-4}$. The total number of epoch is 200 and the epoch milestones are ${50,100,160}$. The learning rate is initialized to $3.5 \times 10^{-5}$ and is decayed by a factor of 0.1 when the epoch get the milestones. At the beginning, we warm up the models for 10 epochs and the learning rate grows linearly from $3.5 \times 10^{-5}$ to $3.5 \times 10^{-4}$. For data sampler, we choose a fully connect layer as energy function. We update the selection probability of each sample with the newest data sampler every $N$ iteration ($N$ is the size of the dataset). 

\subsection{Performance Comparison}
To verify the effectiveness of our model, we compare our method with some of the state-of-art re-sampling and re-weighting methods on Market-1501, DukeMTMC and MSMT-17. 
We further conduct experiments to evaluate the robustness of our method under different types of distribution property. To evaluate our method's robustness on class imbalance, we compare the performance of our methods on dataset with different degree of class imbalance.  To evaluate our methods ability to converge on hard samples (i.e. high intra-identity variation and low inter-identity variation samples), we compare the performance on cross-viewpoint test set. 

\subsubsection{Overall Performance Comparison}
We compare our method with state-of-art re-sampling and re-weighting method on three benchmark datasets. %

The performance comparison of both image sampling methods for cross entropy loss and triplet sampling methods for triplet loss are reported. For a fair comparison, following image sampling methods run under the same experiment setting as our proposed methods: 

\begin{table*}[!t]
% p{3cm}<{\centering}
\centering
\footnotesize
\caption{Performance (\%) comparisons to the state-of-the-art results on Market-1501, DukeMTMC-reID and MSMT-17. $\dag$: result quote from original paper. } 
\label{table_sota_performance}
\resizebox{\textwidth}{!}{
\begin{tabular}{cllccccccccc}
\hline
\multirow{2}*{Category} & \multirow{2}*{Method} & \multicolumn{3}{c}{Market-1501} & \multicolumn{3}{c}{DukeMTMC-reID} & \multicolumn{3}{c}{MSMT-17} \\
\cline{3-11}
& & {mAP}&{Rank-1}&{Rank-5}&{mAP}&{Rank-1}&{Rank-5}&{mAP}&{Rank-1}&{Rank-5}\\
\hline 

\multirow{6}*{ \makecell{Sampling \\ Images}} 
& Baseline (CE)  \cite{Luo_2019_CVPR_Workshops} & 78.73 & 91.98 & 96.85 & 70.12 & 83.89 & 92.46 & 46.75 & 72.25 & 84.09\\
& Data Prior Distribution  \textsuperscript{$\dag$} \cite{wu2018person} & 65.87 & 86.90 & 95.37 & 53.42 & 72.83 & - & - & - & - \\
& Focal Loss \cite{lin2017focal} & 79.03 & 90.50 & 96.20 & 67.76 & 81.46  & 90.84 & 44.32 & 68.91 & 82.58\\
& Self-paced Learning \cite{zhou2018deep}  & 80.61 & 91.78 & 97.27 & 63.07 &77.29 & 88.69 & 50.01 & 74.28 & 85.93 \\ 
& Meta-Weight Net \cite{shu2019meta} & 79.04  & 91.54 & 96.64 & 70.10 & 82.90 & 91.97 & 45.43 & 71.40 & 83.27 \\ 
&  \textbf{Our Methods} & \textbf{82.27} & \textbf{93.02} & \textbf{97.51} & \textbf{73.33} & \textbf{85.86} & \textbf{93.00} &  \textbf{51.60} & \textbf{75.09} & \textbf{86.40}\\
\hline 

\multirow{6}*{\makecell{Sampling \\ Triplets}} 
& Baseline (OHEM) \cite{Luo_2019_CVPR_Workshops}  & 85.7 & 94.1 & - & 75.9 & 86.2 & - & 50.33 & 74.05 & 85.68\\
& Triplet loss OHEM \textsuperscript{$\dag$} \cite{HermansBeyer2017Arxiv}  & 69.14 & 84.92 & 94.21 & - & - & - & - & - & -\\
& MVP Loss \textsuperscript{$\dag$} \cite{sun2019mvp} & 80.5 & 91.4 & - & 70.0 & 83.4 & - & - & - & - \\
& Mancs \textsuperscript{$\dag$} \cite{wang2018mancs} & 82.3 & 93.1 & - & 71.8 & 84.9 & - & - & -   \\
& SemiHard OEM  \cite{schroff2015facenet} & 85.66 & 93.94 & 98.07 & 75.83 & 86.67 & 94.03 & 49.93 & 73.44 & 85.38\\
& \textbf{Our Methods} & \textbf{87.94} & \textbf{94.89} & \textbf{98.16} & \textbf{78.58} & \textbf{88.96} & \textbf{94.75} & \textbf{54.69} & \textbf{78.35} & \textbf{88.40}\\
\hline

\end{tabular}
}
\end{table*}

\begin{table}[!t]
% p{3cm}<{\centering}
\footnotesize
\centering
\caption{Performance (\%) comparisons of the class-balance resist method and our method on Market-1501 with different level of class imbalance. 
%\ancong{/*** ancong: maybe more related re-sampling and re-weighting methods should be compared here ***/}
} 
\label{table_imbalance_performance}
\begin{tabular}{cllcc}
\hline
\multirow{2}*{Noise Level} & \multirow{2}*{Method} & \multicolumn{3}{c}{Market-1501} \\
\cline{3-5}
& & {mAP}&{Rank-1}&{Rank-5}\\
\hline 

\multirow{3}*{90\% few shot} & Baseline & 71.07  & 87.29 & 95.64  \\
& Up-sample & 74.63 & 89.28 & 96.2\\
& This work  & \textbf{81.57} & \textbf{92.70} &  \textbf{97.36}\\ 
\hline 

\multirow{3}*{80\% few shot} 
& Baseline  & 76.95 & 90.38 & 96.79 \\
& Up-sample & 78.44 & 91.06 & 96.97 \\
& This work &  \textbf{83.88} &  \textbf{93.79} & \textbf{97.83} \\
\hline

\multirow{3}*{70\% few shot} &  Baseline & 79.80 & 91.63 & \textbf{97.12} \\
 & Up-sample & 80.50 & 91.75 & 97.51  \\
 & This work & \textbf{84.68} & \textbf{93.56} & \textbf{97.92} \\

\hline
\end{tabular}
\end{table}

\begin{itemize}[leftmargin=*]
    \vspace{-4pt}
    \item \textbf{Strong Baseline w/o triplet loss} \cite{Luo_2019_CVPR_Workshops}. We choose this method as a baseline for image sampling methods. To compare the performance of methods sampling single images, this baseline does not use triplet loss and triplet sampling.  %Following the setting of Strong Baseline with warm-up learning rate schedule and label smoothing regularization but discard the triplet loss connected to the BN neck, this methods use a uniform sampling strategy. 
    \vspace{-4pt}
    \item \textbf{Focal loss}  \cite{lin2017focal}. This is a hard sampling method where the identity loss of each sample is re-weighted with a focal loss weight function. 
    \vspace{-4pt}
    \item \textbf{Self-paced Learning} \cite{zhou2018deep}. This is a method that learns easy sampler where the identity loss of the strong baseline method is replaced with self-paced learning based loss. 
    \vspace{-4pt}
    \item \textbf{Meta-Weight Net} \cite{shu2019meta}. This is a method that learn hard sampler where the identity loss of the samples is re-weighted by a weighting function automatically obtained by meta-learning.
    
\end{itemize}

Following pair sampling methods are compared under the same experiment setting with our proposed methods: 
\begin{itemize}[leftmargin=*]
    \vspace{-4pt}
    \item \textbf{Strong Baseline (OHEM)} \cite{Luo_2019_CVPR_Workshops}. We choose this method as a baseline for triplet sampling methods. 
    \vspace{-4pt}
    \item \textbf{SemiHard OEM}  \cite{schroff2015facenet}. This is a method that select hard samples where the OHEM sampling strategy in baseline is replaced with Semi-Hard sampling strategy.% where triplets whose negative sample's distance from the anchor is smaller than positive sample but within the difference margin are selected. 
\end{itemize}
As the source code is not available for some of the state-of-the-art methods   \cite{wu2018person,wang2018mancs,sun2019mvp}, we quote the results of these methods directly from the original paper, where Mancs \cite{wang2018mancs} is a curriculum learning method to select easy samples, MVP loss \cite{sun2019mvp} is a hard sample selecting method and Data Prior Distribution select samples based on attributes \cite{wu2018person}.  These methods use similar network structure and loss function to produce relatively fair and comparable results to our methods. 
As shown in Table \ref{table_sota_performance}, for both sampling images and triplets, our method outperforms state-of-the-art methods and achieves the best performance, which verifies the advantage of one-for-more based data sampler for ReID. 

We observe that some of the predefined re-weighting/re-sampling methods (e.g, ohem) achieve better performance compared to the uniform sampling baseline, while others do not improve the performance (e.g., semihard ohem). These results show that the predefined re-weighting/re-sampling methods do not generalize on all types of data distribution and extra work is needed to customize selection strategy. Furthermore, some of the methods only achieve performance improvement on one or two of the three datasets and achieve worse performance than baseline on the others. For example, focal loss only boost the performance on Market-1501 and self-paced learning only boost the performance on Market1501 and MSMT-17. This result further proves that predefined strategy does not generalize well on different datasets. 

We observe that our method outperforms the meta-weight net, which also automatically learns a local re-weighting function for cross entropy loss. This results show that leaning global sampling policy and pairwise data sampler instead of re-weighting individual image on local batch is more suitable for person ReID task and effectively boosts the performance.

\subsubsection{Performance on Imbalanced Dataset}

To verify the effectiveness of our method on imbalanced dataset, we manually enhance the level of identity balance of the existing dataset. For a public ReID dataset, we first sort the identities of the existing dataset based the number of images in them. Then, we down-sample the top-$m$ identities to $n$ samples each ($n = 5$ in our experiments).  Table \ref{table_imbalance_performance} shows the performance of baseline methods and our method on imbalance dataset with different level of identity imbalance. We choose a classic up-sampling strategy as the comparison method, which has proven to be a effective method for imbalanced dataset. We observe that, when there is a large amount of tail data, our method significant outperforms the baseline and existing up-sampling strategy. 

\subsubsection{Evaluation on Hard Samples}
To verify our method's ability to converge on datasets with high intra-identity variation and inter-identity variation, we evaluate our method on a testset with such distribution. Specifically, we annotate the viewpoint of samples on  Market-1501 with three categories: \textit{front, side, back}. For each query in each test case, to increase the inner-class variation, we only select the images with different views as its corresponding gallery image. The performance of our method and the baseline methods on the hard case test set is show in Table \ref{table_view}. We choose hard mining methods as the comparison baseline, which is a classic and state-of-the-art strategy to boost model generalization ability on hard cases. We observe that our method outperforms baseline and hard sample mining method, which verifies that the learned sampling policy help ReID model to generalize better on samples with high inner-class variation, leading to a better performance on  test set with the same distribution. 

\begin{table}[!t]
% p{3cm}<{\centering}
\centering
\footnotesize
\caption{Performance Comparison (\%) of baseline method and our method on cross-viewpoint retrieval test set.} 
\label{table_view}
\begin{tabular}{ccccc}
\hline
Query & Method & {mAP} & {Rank-1}& {Rank-5}\\
\hline
\multirow{2}*{${\textbf{D}_{front}}$} & baseline & 81.49 & 89.82 & 95.99\\
&  Our method & \textbf{84.82} & \textbf{91.50} & \textbf{96.63} \\
\hline  
\multirow{2}*{${\textbf{D}_{side}}$} & baseline & 82.17 & 90.56 & 95.52 \\
& Our method & \textbf{84.47} & \textbf{91.10} & \textbf{96.33} \\
 \hline
\multirow{2}*{${\textbf{D}_{back}}$} & baseline & 82.82 & 91.42 & 97.03 \\
& Our method & \textbf{84.82} & \textbf{93.73} & \textbf{97.03}\\
\hline
\end{tabular}
\end{table}

\begin{figure}[t]
\centering
\includegraphics[width=0.4\textwidth]{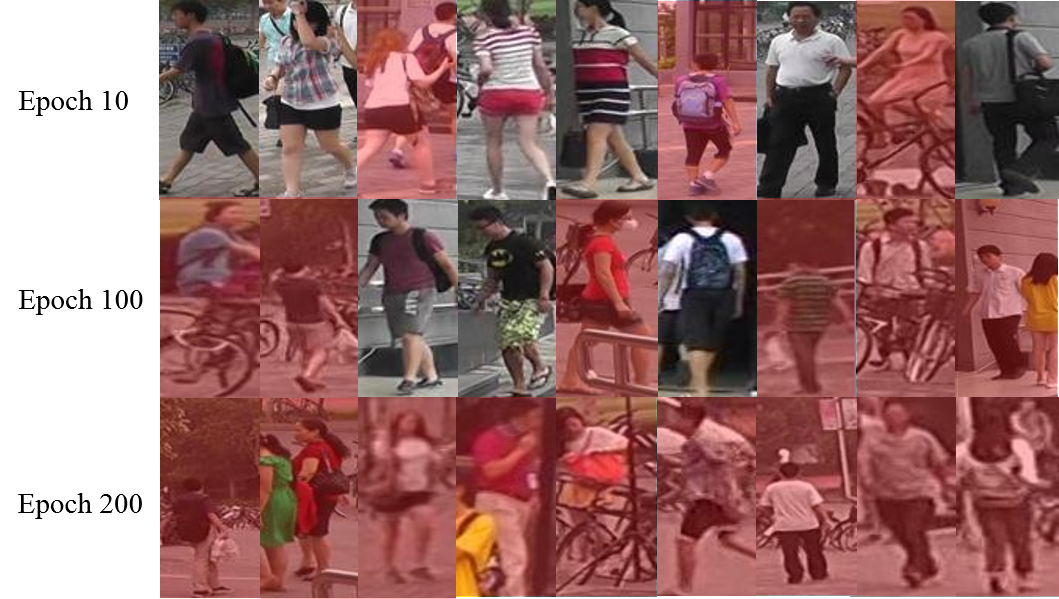} 
\caption{High probability samples in generalized data distribution at different training stages} 
\label{fig_hard_sample}
\end{figure}

%\subsubsection{Domain Adaptation}

To further demonstrate how our method help model learn better on hard samples, Figure \ref{fig_hard_sample} visualize the images with high probability to be selected by sampler at different training epoch. We observe that the sampling policy changes as the learning processes progresses. At the early stage of the training (i.e. epoch 10 in Figure \ref{fig_hard_sample}), the high probability samples are high-resolution, clean images without strong pose variation and obvious occlusion. We believe this is because at the early stage of training when the model is not sufficiently trained, data sampler tends to select easier samples to help the model quickly converge. On the other hand, at the late stage of the training (epoch 100 / 200 in Figure \ref{fig_hard_sample}), we observe that more and more hard samples are selected (highlighted with red rectangle). We believe this is because when the model is almost converged, it has already sufficiently trained on the easy samples, so the data sampler should lower the importance of the easy samples to avoid over-fitting on them, and pay more attention to the under-fit hard samples.

\begin{table}[t]
% p{3cm}<{\centering}
\centering
\footnotesize
\caption{Performance (\%) comparisons of the noise resist method and our method on Market-1501 with different level of label noise added. } 
\label{table_noisy_performance}
\begin{tabular}{cllcc}
\hline
\multirow{2}*{Noise Level} & \multirow{2}*{Method} & \multicolumn{3}{c}{Market-1501} \\
\cline{3-5}
& & {mAP}&{Rank-1}&{Rank-5}\\
\hline 

\multirow{3}*{5\% label switched} & Baseline & 75.75 & 89.85 & 96.64 \\
& PENCIL & 71.77 & 87.50 & 95.40  \\
& \textbf{This work}  &   \textbf{79.65}   &  \textbf{92.22} & \textbf{97.12} \\ 
\hline 

\multirow{3}*{10\% label switched}
& Baseline & 64.72 & 83.88 & 94.24 \\
& PENCIL & 68.24 & 86.16 & 94.92 \\
&  \textbf{This work} &  \textbf{71.05} &  \textbf{88.24} & \textbf{97.12} \\
\hline

\multirow{3}*{15\% label switched} &  Baseline & 59.54 & 80.14 & 93.35  \\
& PENCIL & 64.92 & 83.37 & 93.91 \\
 & This work &  62.61  & 83.19  & 94.15  \\

\hline
\end{tabular}
\end{table}

\subsubsection{Performance on Noisy Dataset}
We evaluate the influence of noisy labels on our method. Table \ref{table_noisy_performance} shows the performance of baseline methods and our method on the noisy dataset with different percentage of noisy label added. We choose PENCIL \cite{yi2019probabilistic}, which is one of the state-of-the-art noise resist methods. We observe that, when the number of noisy label is low ($5\%$ and $10\%$ ), our method outperforms the baseline method and SOTA noise resist method. However when the level of noisiness progress ($15\%$), our method is not able to perform as well as PENCIL. The experiments results show that our method can resist certain level of noise in the training set, which is consistent with the real-world ReID data distribution.  This is because when the level of noisiness is low, the data sampler is robust enough to select clean samples to help ReID model converge, but when the noisiness level is high, it may make ReID model fit over-fit on both clean samples and the noisy samples. 

\begin{table}[t]
% p{3cm}<{\centering}
\centering
\footnotesize
\caption{Performance Comparison (\%) of baseline method and our method on CIFAR10.} 
\label{table_cifar_performance}
\begin{tabular}{cc}
\hline
Method & {Accuracy} \\
\hline
Random Sampling \cite{thangarasa2018self} & 93.26 \\
Focal Loss & 92.59 \\
Self-paced learning \cite{thangarasa2018self} & 94.31 \\
Meta-weight Net \cite{shu2019meta} & 92.4 \\ 
Our method & \textbf{96.84} \\
\hline
\end{tabular}
\end{table}
\subsubsection{Performance on other Task}
To demonstrate our method's effectiveness on the task other than person ReID, we compare our learned data sampler with existing re-sampling methods on classification task with  CIFAR10 dataset \cite{krizhevsky2009learning}. Table \ref{table_cifar_performance} shows the accuracy comparison of our methods and some state-of-the-art re-sampling methods and our method achieves the highest performance. 

\section{Conclusion}
This paper proposes an one-for-more objective to learn a data sampler that selects generalizable training sample to train model towards generalize on all data. The data sampler learning is integrated with ReID model learning,  forming an end-to-end unified training framework.  The experiments show that our model outperforms the state-of-the-art online re-sampling and re-weighting methods and show impressive generalization ability on different types of distributions. 

% ---- Bibliography ----
%
% BibTeX users should specify bibliography style 'splncs04'.
% References will then be sorted and formatted in the correct style.
%
%\bibliographystyle{aaai21}
\bibliography{egbib}

\end{document}